\documentclass[10pt,twocolumn,letterpaper]{article}

\usepackage{iccv}
\usepackage{times}
\usepackage{epsfig}
\usepackage{graphicx}
\usepackage{subcaption}
\usepackage{amsmath}
\usepackage{amssymb}
\usepackage{bm}
\usepackage{multicol}
\usepackage{multirow}
\usepackage{tabu}
\usepackage{algorithm,algcompatible}
\algnewcommand\INPUT{\item[\textbf{Input:}]}
\algnewcommand\OUTPUT{\item[\textbf{Output:}]}
\usepackage{xcolor}
\usepackage{caption}
\usepackage[font=small]{caption}
\makeatletter
\def\thickhline{%
  \noalign{\ifnum0=`}\fi\hrule \@height \thickarrayrulewidth \futurelet
   \reserved@a\@xthickhline}
\def\@xthickhline{\ifx\reserved@a\thickhline
               \vskip\doublerulesep
               \vskip-\thickarrayrulewidth
             \fi
      \ifnum0=`{\fi}}
\makeatother

\newlength{\thickarrayrulewidth}
\newcolumntype{?}{!{\vrule width 1pt}}

\usepackage[pagebackref=true,breaklinks=true,letterpaper=true,colorlinks,bookmarks=false]{hyperref}

\iccvfinalcopy % *** Uncomment this line for the final submission

\def\iccvPaperID{12} % *** Enter the CVPR Paper ID here

\ificcvfinal\fi

\begin{document}

%%%%%%%%% TITLE
\title{Cross Domain Image Matching in Presence of Outliers}

\author{Xin Liu\ \ \ \ Seyran Khademi\ \ \ \ Jan C. van Gemert\\
Computer Vision Lab, Delft University of Technology\\
Delft, The Netherlands
%{\tt\small x.liu-11@tudelft.nl}
}

\maketitle
\ificcvfinal\thispagestyle{empty}\fi

%%%%%%%%% ABSTRACT
\begin{abstract}
   Cross domain image matching between  image collections from different source and target domains is challenging in times of deep learning due to i) limited variation of  image conditions in a training set, ii) lack of paired-image labels during training, iii) the existing of outliers that makes image matching domains not fully overlap.
   To this end, we propose an end-to-end architecture that can match cross domain images without labels in the target domain and handle non-overlapping domains by outlier detection.
   We leverage domain adaptation and triplet constraints for training a network capable of learning domain invariant and identity distinguishable representations, and iteratively detecting the outliers with an entropy loss and our proposed weighted MK-MMD.
   Extensive experimental evidence on \textit{Office}~\cite{Saenko:2010:AVC:1888089.1888106} dataset and our proposed datasets \textit{Shape, Pitts-CycleGAN} shows that the proposed approach yields state-of-the-art cross domain image matching and outlier detection performance on different benchmarks.
   The code will be made publicly available.
   
\end{abstract}

%%%%%%%%% BODY TEXT
\section{Introduction} 
\label{sec:intro}
Cross domain image matching is about matching two images that are collected from different sources (\eg photos of the same location but captured in different illuminations, seasons or era).  
It has wide application value in different areas, with research in location recognition over large time lags~\cite{Fernando2015timelag}, e-commerce product image retrieval~\cite{Ji2017attentionmodeling}, urban environment image matching for geo-localization~\cite{Tian2017geolocalization}, etc.

\begin{figure}[t]
\centering
\begin{subfigure}{0.235\textwidth}
\centering
\includegraphics[width=1.0\linewidth]{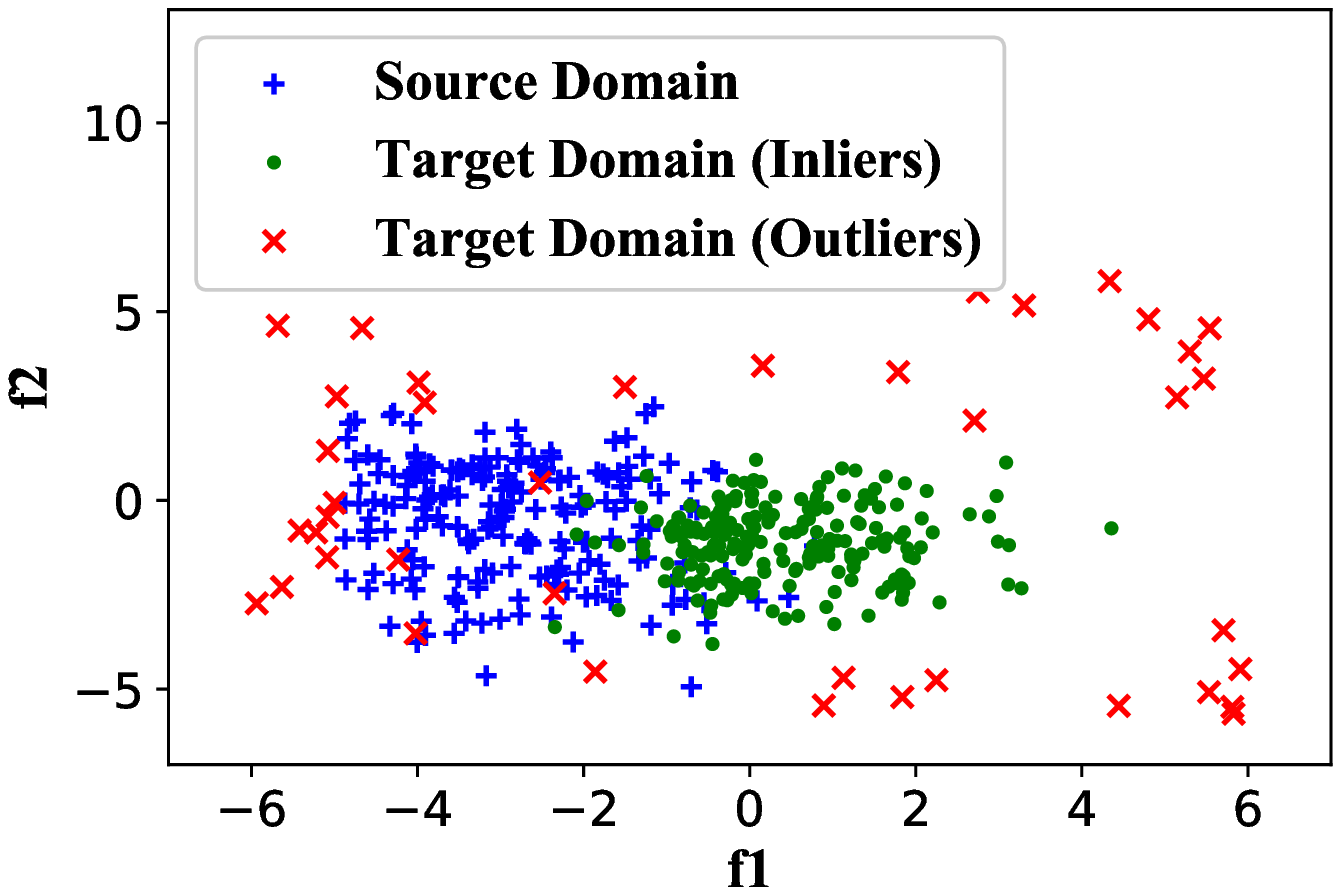}
\caption{Original sample distribution}
\end{subfigure}%
\begin{subfigure}{0.24\textwidth}
\centering
\includegraphics[width=1.0\linewidth]{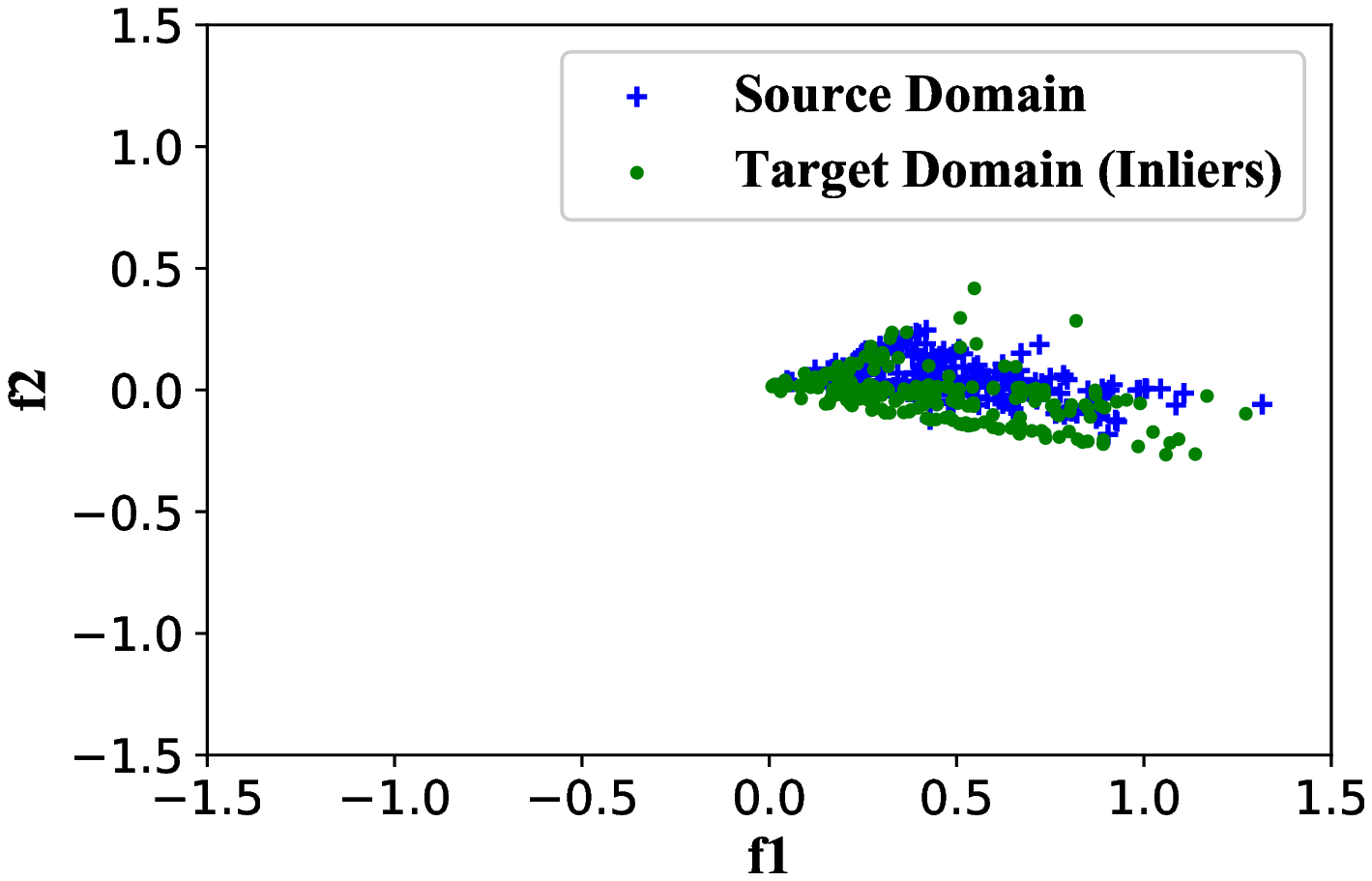}
\caption{Matching + DA}
\end{subfigure}
\begin{subfigure}{0.24\textwidth}
\centering
\includegraphics[width=1.0\linewidth]{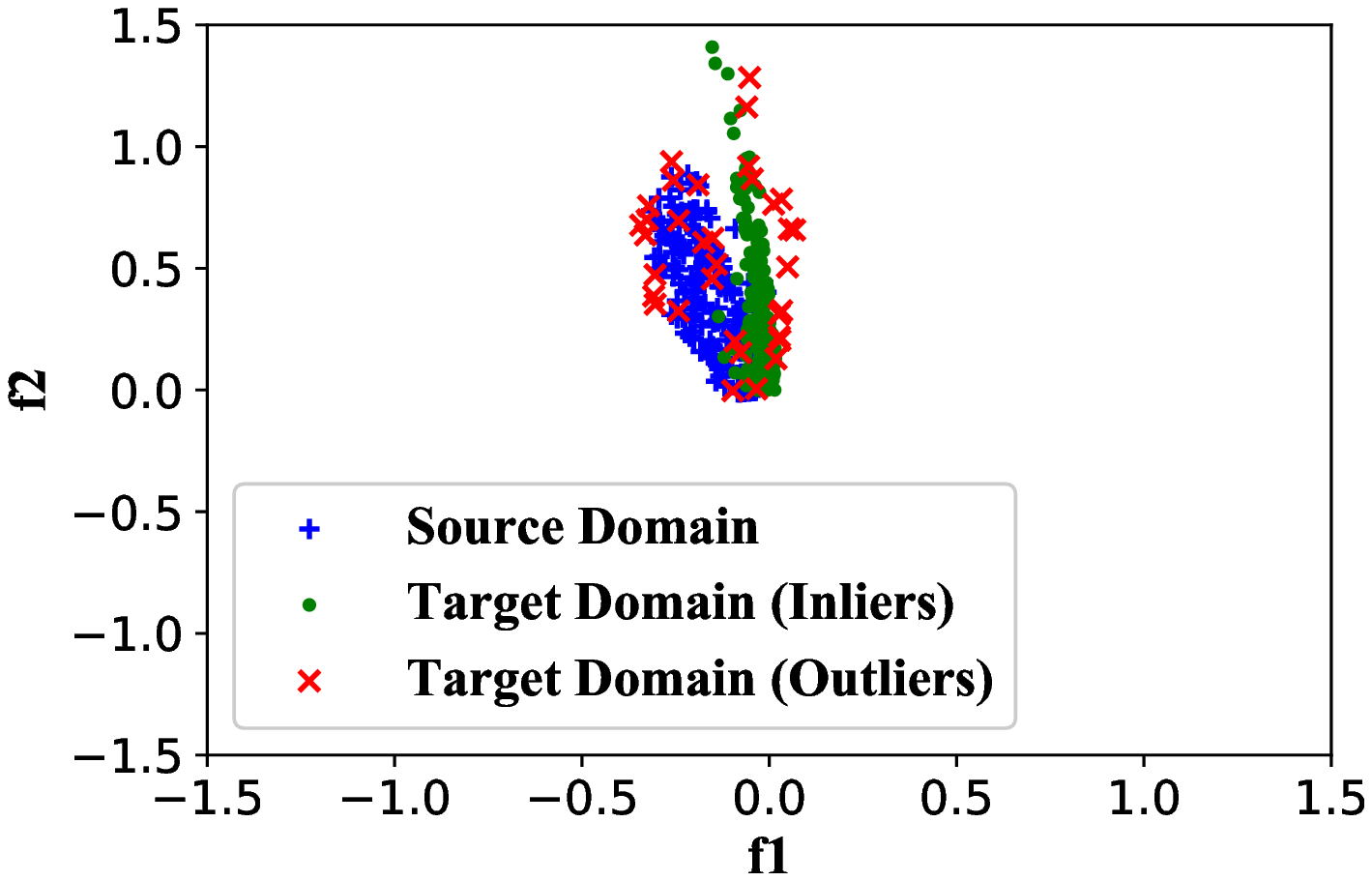}
\caption{Matching + DA}
\end{subfigure}%
\begin{subfigure}{0.24\textwidth}
\centering
\includegraphics[width=1.0\linewidth]{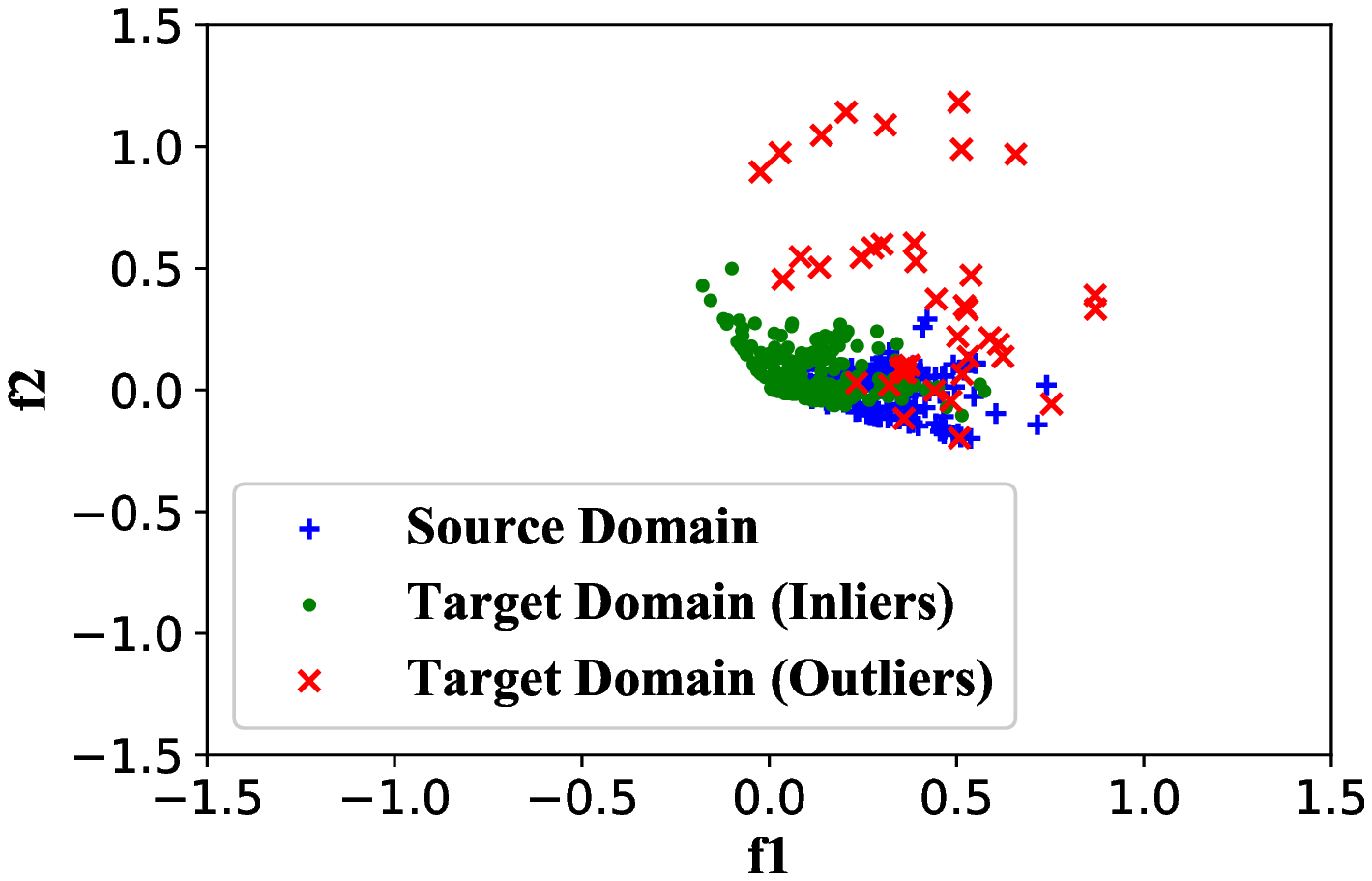}
\caption{Our method}
\end{subfigure}
\caption{Domain adaptation (DA) and image matching applied on a 2D toy dataset generated with domain shift between source and target domains. (a) Original distribution, (b) no outliers, (c) with outliers, (d) our method. The result of (b) and (c) shows that outliers affect the alignment of source samples and inlier target samples. (c) and (d) show that our outlier detection helps separating the outliers from the aligned source samples and inlier target samples.}
\label{fig:concept}
\end{figure}
% problem
Even using deep feature representation learning, the automated cross domain image matching task remains challenging mainly due to the following difficulties.
First, it is difficult to match varying observations of the same location or object, in general. 
Second, often the paired-image examples from two domains are not available for training neural networks.
Third, the image samples in two domains may not fully overlap due to the existing of outlier images, which affects the matching performance if such outliers are not detected.

% method in a nutshell
In this work, we address the problem of domain adaptation for  feature learning in a cross domain matching task when outliers are present. 
As is common in domain adaptation, we only have labeled image pairs from the source domain, but no labels from the target  domain.
To resolve the domain disparity between the train and the test data, we are inspired from Siamese network~\cite{Chopra2005siamse} for image matching and  domain adaptation used in image classification~\cite{Long2015DAN,Saito17asymmetrictri-training,Tzeng2017ADDA,Venkateswara2017DHN,Zhang2015DTN}. 
We propose a triplet constraints network to learn the domain invariant and identity distinguishable representations of the samples.
This is made possible by utilizing the paired-image information from the source domain, a weighted multi-kernel maximum mean discrepancy (weighted MK-MMD) method and an entropy loss.
The setting of the problem and experiment results of our method are depicted on a 2D toy dataset in Figure \ref{fig:concept}.

% contribution
To verify our method, we introduce two new synthetic datasets as there are no publicly available datasets for our problem setting. Moreover, we believe outlier-aware algorithms are essential to design practical domain adaptation algorithms as many real data repositories contain irrelevant samples w.r.t. the source domain.  
In summary, our main contribution is two-fold:
\begin{itemize}
\item Joint domain adaptation and outlier detection.
\item Two new datasets, \textit{Pits-CycleGAN} dataset and \textit{Shape} dataset, for cross domain image matching. 
\end{itemize}

\section{Related work}
\label{sec:related}
\subsection{Image matching}
Feature learning based matching methods become popular due to its improved performance over hand-crafted features (\eg SIFT~\cite{Lowe1999SIFT}). 
Siamese network architectures~\cite{Chopra2005siamse} are among the most popular feature learning networks, especially for pairs comparison tasks. We also adopt Siamese network as part of our framework.
The purpose is to learn feature representations to distinguish matching and unmatching pairs in the source domain, which assists the network in learning to match cross domain images.
In the cross-domain image matching context, Lin \etal~\cite{Lin_2015_CVPR} investigated a deep Siamese network to learn feature embedding for cross-view image geo-localization.
Kong \etal~\cite{Kong2018deepfeaturemaps} applied Siamese architecture to cross domain footprint matching. 
Tian \etal~\cite{Tian2017geolocalization} utilized Siamese network for matching the building images from street view and bird's eye view.
Unlike the existing works on cross-domain image matching, we consider labeled paired-image information is only available in the source domain.

% In addition, we also consider the outliers existing in the query target domain.
% To solve this problem, we propose to apply domain adaptation methods in our unsupervised cross domain image matching and perform outliers detection at the same time. 
\subsection{Domain adaptation}
Domain adaptation have been researched over recent years in  diverse domain classification tasks, in which adversarial learning and statistic methods are main approaches. 
Ganin \etal~\cite{Ganin2016DAT} proposed domain-adversarial training of neural networks with input of labeled source domain data and unlabeled target domain data for classification.
% Tzeng \etal~\cite{Tzeng2017ADDA} proposed a framework which combines a GAN~\cite{Goodfellow2014GAN} loss for unsupervised adaptation classification. 
In~\cite{Zhang2015DTN}, the authors proposed a deep transfer network (DTN), which achieved domain transfer by simultaneously matching both the marginal and the conditional distributions with adopting the empirical maximum mean discrepancy (MMD)~\cite{Gretton2006KMT}, which is a nonparametric metric.
Venkateswara \etal~\cite{Venkateswara2017DHN} applied MK-MMD~\cite{NIPS2012_4727} to a deep learning framework that can learn hash codes for domain adaptive classification. In this setting MK-MMD loss promotes  nonlinear alignment of data, which generates a nonparametric distance in Reproducing Kernel Hilbert Space (RKHS).
The distance between two distributions is the distance between their means in a RKHS.
When two data sets belong to the same distribution, their MK-MMD is zero.
Based on the successful performance of MK-MMD loss, we also adopt it to adapt different domains, this time for image matching task. This requires the marriage of Siamese network with MK-MMD loss, as we do  later in our paper.

\subsection{Outlier detection} 
Much work exists on outlier detection~\cite{Chalapathy2018oneclassneuralnetworks,Liu2014UOL, Sabokrou2018adversariallyone-class,Zhang_2018_CVPR}. Chalapathy \etal~\cite{Chalapathy2018oneclassneuralnetworks} proposed an one-class neural network (OC-NN) encoder-decoder model to detect anomalies.
Sabokrou \etal~\cite{Sabokrou2018adversariallyone-class} also applied the encoder-decoder architecture as part of their network for novelty detection.
Zhang \etal~\cite{Zhang_2018_CVPR} proposed an adversarial network for partial domain adaptation to deal with outlier classes in the source domain. 
Their network is for classification task, and they do not have the assumption that outliers originate from low-density distribution.
Instead, we are inspired by the work of Liu \etal~\cite{Liu2014UOL} which uses a kernel-based method to learn, jointly, a large margin one-class classifier and a soft label assignment for inliers and outliers.
Using the soft label assignment, we implement outlier detection with  cross domain image matching in an iterative sample reweighting way.

\begin{figure*}[t]
\centering
\includegraphics[width=0.6\textwidth]{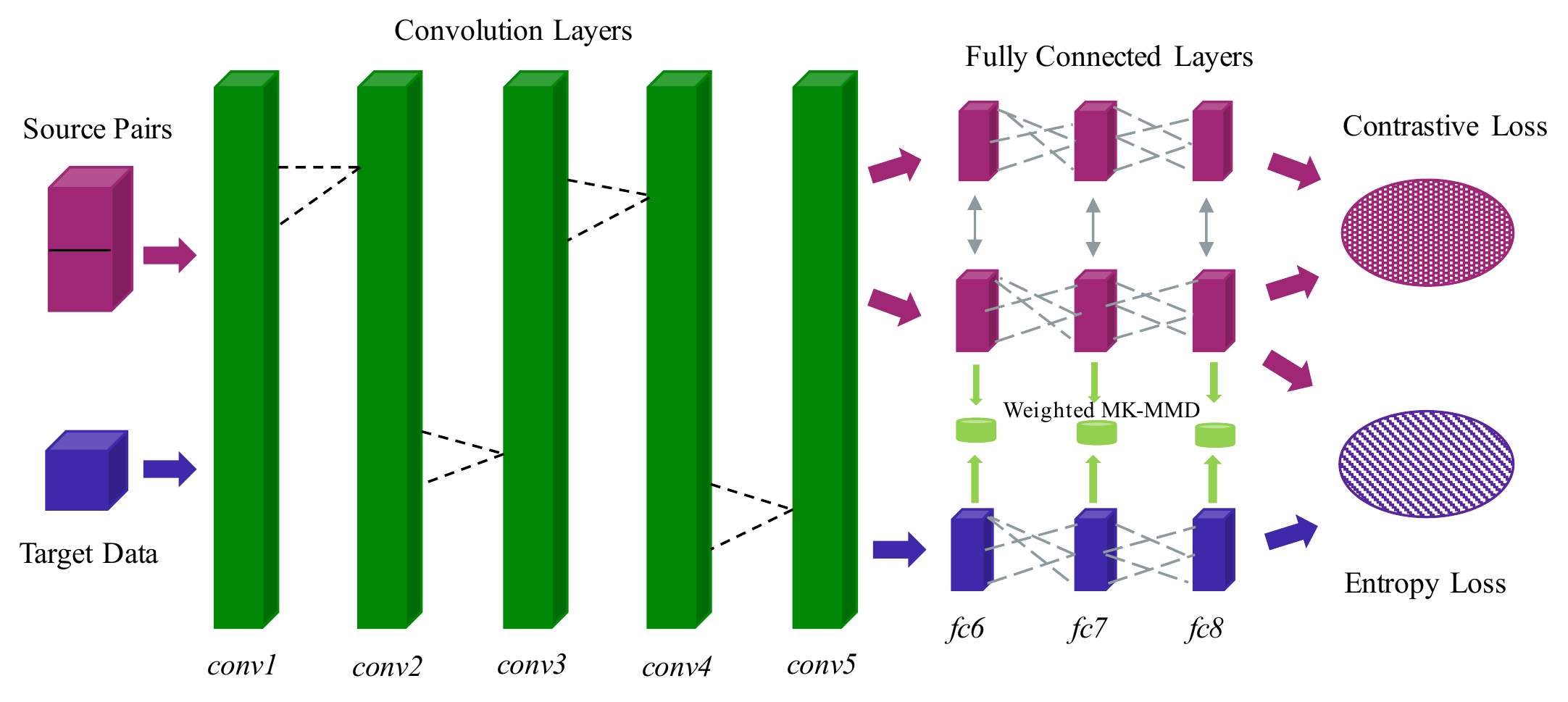}
\caption{The network for  cross domain image matching and outlier detection.
The contrastive loss makes the network to learn paired-image information from the source.
The weighted MK-MMD loss trains the network to learn transferable features between the source and the inliers of the target.
The entropy loss helps distinguish inliers and outliers in the target domain.
}
\label{fig:network}
\end{figure*}

\section{Domain adaptive image matching}

\subsection{Siamese loss}
We introduce our proposal for domain adaptation for image matching task once labeled data is not available in the target domain. 
Let $X_{s}$ denote the source domain image set. 
A pair of images $x_{i}, x_{j} \in X_{s}$ are used as input to part of our network, as shown in Figure \ref{fig:network}.
$x_{i}, x_{j}$ can be a matching pair or an unmatching pair.
The objective is to automatically learn a feature representation, $f(\cdot)$, that effectively maps the input $x_{i},x_{j}$ to a feature space, in which matching pairs are close to each other and unmatching pairs are far apart.
We employ the contrastive loss as introduced in ~\cite{Hadsell:2006:DRL:1153171.1153654}:
\begin{equation}
L(x_{i},x_{j},y)=\frac{1}{2}yD^{2} + \frac{1}{2}(1-y)\{\max(0,m-D)\}^{2},
\label{eq:1}
\end{equation}
where $y\in \{0,1\}$ indicates unmatching pairs with $y=0$ and matching pairs with $y=1$, $D$ is the Euclidean distance between the two feature vectors $f(x_{i})$ and $f(x_{j})$, and $m$ is the margin parameter acting as threshold to separate matching and unmatching pairs.

\subsection{Domain adaptation loss}
\label{sub:3.2}
It is known that in deep CNNs, the feature representations transition from generic to task-specific as one goes up from bottom layers to other layers~\cite{Yosinski:2014:TFD:2969033.2969197}. 
Compared to the convolution layers \textit{conv1} to \textit{conv5}, the fully connected layers are more task-specific and need to be adapted before they can be transferred~\cite{Venkateswara2017DHN}.

Accordingly, our approach attempts to minimize the MK-MMD loss to reduce the domain disparity between the source and target feature representations for fully connected layers, $\mathcal{F}=\{fc6,fc7,fc8\}$.
The multi-layer MK-MMD loss is given by,
\begin{equation}
\label{eq:2}
\mathcal{M}(u_{s},u_{t})=\sum_{l\in \mathcal{F}}d_{k}^{2}(u_{s}^{l},u_{t}^{l}),
\end{equation}
where, $u_{s}^{l}=\{\bm{u}_{i}^{s,l}\}_{i=1}^{n_{s}}$ and $u_{t}^{l}=\{\bm{u}_{i}^{t,l}\}_{i=1}^{n_{t}}$ are the set of output representations for the source and target data at layer $l$, $\bm{u}_{i}^{*,l}$ is the output representation of inuput image $\bm{x}_{i}^{*,l}$ for the $l^{th}$ layer.
The MK-MMD measure $d_{k}^{2}(\cdot)$ is the multi-kernel maximum mean discrepancy between the source and target representations~\cite{NIPS2012_4727}.
For a nonlinear mapping $\phi(\cdot)$ associated with a reproducing kernel Hilbert space $\mathcal{H}_{k}$ and kernel $k(\cdot)$, where $k(\bm{x},\bm{y})=\langle\phi(\bm{x},\bm{y})\rangle$, the MK-MMD is defined as,
\begin{equation}
\label{eq:3}
d_{k}^{2}(u_{s}^{l},u_{t}^{l})=||\mathrm{E}[\phi(\bm{u}^{s,l})]-\mathrm{E}[\phi(\bm{u}^{t,l})]||_{\mathcal{H}_{k}}.
\end{equation}
The characteristic kernel $k(\cdot)$, is determined as a convex combination of $\kappa$ PSD kernels, $\{k_{m}\}_{m=1}^{\kappa}$, $K := \{k: k=\sum_{m=1}^{\kappa}\beta_{m}k_{m}, \sum_{m=1}^{\kappa}\beta_{m}=1, \beta_{m}\geq0, \forall m\}$.
In particular, we follow~\cite{Long:2016:UDA:3157096.3157112} and set the kernel weights as $\beta_{m}=1/\kappa$ .

\section{Proposed method: Outlier-aware domain adaptive matching}
\label{sec:approach}
% problem definition
The task is to match images with the same content but from different domains where the outliers are present in the target domain. We assume that in the source domain there are sufficient labeled image pairs and in the target domain low-density outliers are present. As in conventional domain adaptation setting labeled data is not available in the target domain.  
We propose a deep triplet network which is comprised of three instances of the same feed-forward network with shared parameters, as shown in Figure \ref{fig:network}.
\subsection{Importance weighted domain adaptation}
\label{sub:4.2}
In our implementation, the MK-MMD loss in subsection \ref{sub:3.2} is calculated over every batch of data points during the back-propagation.
Let n (even) be the number of source data points $u_{s} := \{\bm{u}_{i}^{s}\}_{i=1}^{n}$ and the number of target data points $u_{t} := \{\bm{u}_{i}^{t}\}_{i=1}^{n}$ in the batch.
Then, the MK-MMD can be defined over a set of 4 data points $\bm{z}_{i}=[\bm{u}_{2i-1}^{s},\bm{u}_{2i}^{s},\bm{u}_{2i-1}^{t},\bm{u}_{2i}^{t}],\ \forall i \in \{1,2,...,n/2\}$.
Thus, the MK-MMD is given by,
\begin{equation}
\label{eq:4}
d_{k}^{2}(u_{s},u_{t})=\sum_{m=1}^{\kappa}\beta_{m}\frac{1}{n/2}\sum_{i=1}^{n/2}h_{m}(\bm{z}_{i}),
\end{equation}
where, $\kappa$ is the number of kernels and $\beta_{m}=1/\kappa$ is the weight for each kernel.
And we can expand $h_{m}(\cdot)$ as,
\begin{multline}
\label{eq:5}
h_{m}(\bm{z}_{i}) = k_{m}(\bm{u}_{2i-1}^{s}, \bm{u}_{2i}^{s}) + k_{m}(\bm{u}_{2i-1}^{t}, \bm{u}_{2i}^{t}) \\
- k_{m}(\bm{u}_{2i-1}^{s}, \bm{u}_{2i}^{t}) - k_{m}(\bm{u}_{2i}^{s}, \bm{u}_{2i-1}^{t}),
\end{multline}
in which, the kernel is $k_{m}(\bm{x},\bm{y}) = \exp(-\frac{||\bm{x}-\bm{y}||_{2}^{2}}{\sigma_{m}})$.

With equations \ref{eq:4} and \ref{eq:5}, we can interpret that in the minimum calculation unit ($h_{m}(z_{i})$),  two target domain images contribute to MK-MMD loss calculation.
When there are outliers in the target domain, we only want the inliers to contribute to the calculation, but not the outliers.
Therefore, we could assign the target samples with weights $w_{i}$ as 1 for inliers, and 0 for outliers.
Because we have no ground truth labels, we can only treat the weights as the probability of the target samples to be inliers.
Hence, we can introduce the weighted MK-MMD as,
\begin{equation}
\label{eq:6}
d_{w_k}^{2}(u_{s},u_{t})=\sum_{m=1}^{\kappa}\beta_{m}\frac{1}{n/2}\sum_{i=1}^{n/2}w_{2i-1}w_{2i}h_{m}(\bm{z}_{i}),
\end{equation}
where, $w_{2i-1}$ and $w_{2i}$ are the weights of the target data points $\bm{u}_{2i-1}^{t}$ and $\bm{u}_{2i}^{t}$ in $h_{m}(z_{i})$ respectively, and $w_{2i-1}, w_{2i} \in [0,1]$.
We will explain how to obtain the weight for each target domain sample in next subsection.
\subsection{Outlier detection}
\label{sub:4.3}
Since the inlier-outlier label is not available, we implement an entropy loss to iteratively reassign target domain sample probability of being an inlier, which provides the weights for the weighted MK-MMD.

We use the similarity measure $\langle\bm{u}_{i}, \bm{u}_{j}\rangle$ to learn discriminative inlier-outlier information for the target domain data.
We define three classes of reference data $u_{r}$ for similarity measure, the source domain class $\bm{u}^{1}$, the pseudo inlier class $\bm{u}^{2}$ and the pseudo outlier class $\bm{u}^{3}$.
An ideal target output $\bm{u}_{i}^{t}$ needs to be similar to many of the outputs from one of the classes, $\{\bm{u}_{k}^{c}\}_{k=1}^{K}$.
We assume $K$ data points for every class $c$, where $c \in \{1,2,3\}$ and $\bm{u}_{k}^{c}$ is the $k^{th}$ output from class $c$.
Then the probability measure for each target sample can be outlined as,
\begin{equation}
\label{eq:7}
p_{ic} = \frac{\sum_{k=1}^{K}\exp({\bm{u}_{i}^{t}}^\intercal \bm{u}_{k}^{c})}{\sum_{c=1}^{C}\sum_{k=1}^{K}\exp({\bm{u}_{i}^{t}}^\intercal\bm{u}_{k}^{c})},
\end{equation}
where, $p_{ic}$ is the probability that a target domain sample $x_{i}^{t}$ is assigned to category $c$.
When the sample output is similar to one category only, the probability vector $\bm{p}_{i}=[p_{i1},...,p_{ic}]^\top$ tends to be a one-hot vector.
A one-hot vector can be viewed as a low entropy realization of $\bm{p}_{i}$.
Thus, we introduce a loss to capture the entropy of the probability vectors.
The entropy loss can be given by,
\begin{equation}
\label{eq:8}
S(u_{r},u_{t})=-\frac{1}{n_{t}}\sum_{i=1}^{n_{t}}\sum_{c=1}^{C}p_{ic}log(p_{ic}).
\end{equation}

In subsection \ref{sub:4.2}, we discussed the weighted MK-MMD loss with weights $w_{2i-1}$ and $w_{2i}$. 
With the sample probabilities of target domain data calculated from equation \ref{eq:7}, the weights are calculated as,
\begin{equation}
\label{eq:9}
w_{i} = 
    \begin{cases}
        \frac{p_{i1}+p_{i2}}{p_{i1}+p_{i2}+p_{i3}} & \text{if $x_{i}^{t}$ is classified as source}\\
        \frac{p_{i2}}{p_{i1}+p_{i2}+p_{i3}} & \text{if $x_{i}^{t}$ is classified as others}
    \end{cases}.
\end{equation}
If a target domain sample is classified as "source", then it has a high probability of being an inlier, and therefore should contribute more to reducing the domain disparity.
So we calculate the weight of such a target domain sample with the sum of $p_{i1}$ and $p_{i2}$.

\paragraph{Algorithm}
We iteratively update the target domain data weights after each epoch during training, which works together with domain adaptation for guiding and correcting the detection of outliers and inliers.

The proposed algorithm for outlier detection is showed in the following.
\begin{algorithm}
    \caption{}
  \begin{algorithmic}[1]
  \label{alg:the_alg}
    \INPUT source domain and target domain training data
    \OUTPUT target domain training data probabilities
    \STATE \textbf{Initialization} $i=0$, calculate the average Euclidean distance of each target domain training sample between all the source domain training samples, sort the distances in ascending order and initialize target domain training samples' weights according to the sorted distances, $x_{i} \in$ first half: $w_{i}=0.7$ (pseudo inlier class), $x_{i} \in$ second half: $w_{i}=0.3$ (pseudo outlier class). Inlier class consists of source domain training data, which has the same number of samples with pseudo inlier and pseudo outlier classes.
    \STATE \textbf{Repeat}:
      \STATE $i=i+1$
      \STATE make new mini batches
      \STATE minimize the overall loss function objective (\ref{eq:10})
      \STATE update the samples' weights by equation \ref{eq:7} and \ref{eq:9}
      \STATE update the sets of pseudo inlier class and pseudo outlier class
    \STATE \textbf{Until} target samples' probabilities are unchanged or training time ends
  \end{algorithmic}
\end{algorithm}
The proposed method is built upon the intuitive assumption that outliers originate from low-density distribution.
Thus, we can assume that the ratio of outliers to all the target domain data is no more than 50\%.

\subsection{Overall objective}
\label{sub:4.4}
We propose a model for  cross domain image matching and outlier detection, which incorporates learning image matching information from source domain (\ref{eq:1}),  weighted domain adaptation between the source and the target (\ref{eq:6}) and outlier detection (\ref{eq:8}) in a deep CNN.
The overall objective is given by:
\begin{equation}
\label{eq:10}
min_{u}\textit{J} = \textit{L}(u_{s}) + \gamma\textit{M}_{w}(u_{s},u_{t}) + \eta \textit{S}(u_{r},u_{t}),
\end{equation}
where, $u := \{u_{s} \bigcup u_{t}\}$ and $(\gamma,\eta)$ control the importance of domain adaptation (\ref{eq:6}) and entropy loss (\ref{eq:8}) respectively.

\section{Experiments}
\label{sec:experiment}
\subsection{Datasets}
\label{sub:5.1}
\begin{figure}[h!]
\centering
   \includegraphics[width=0.9\linewidth]{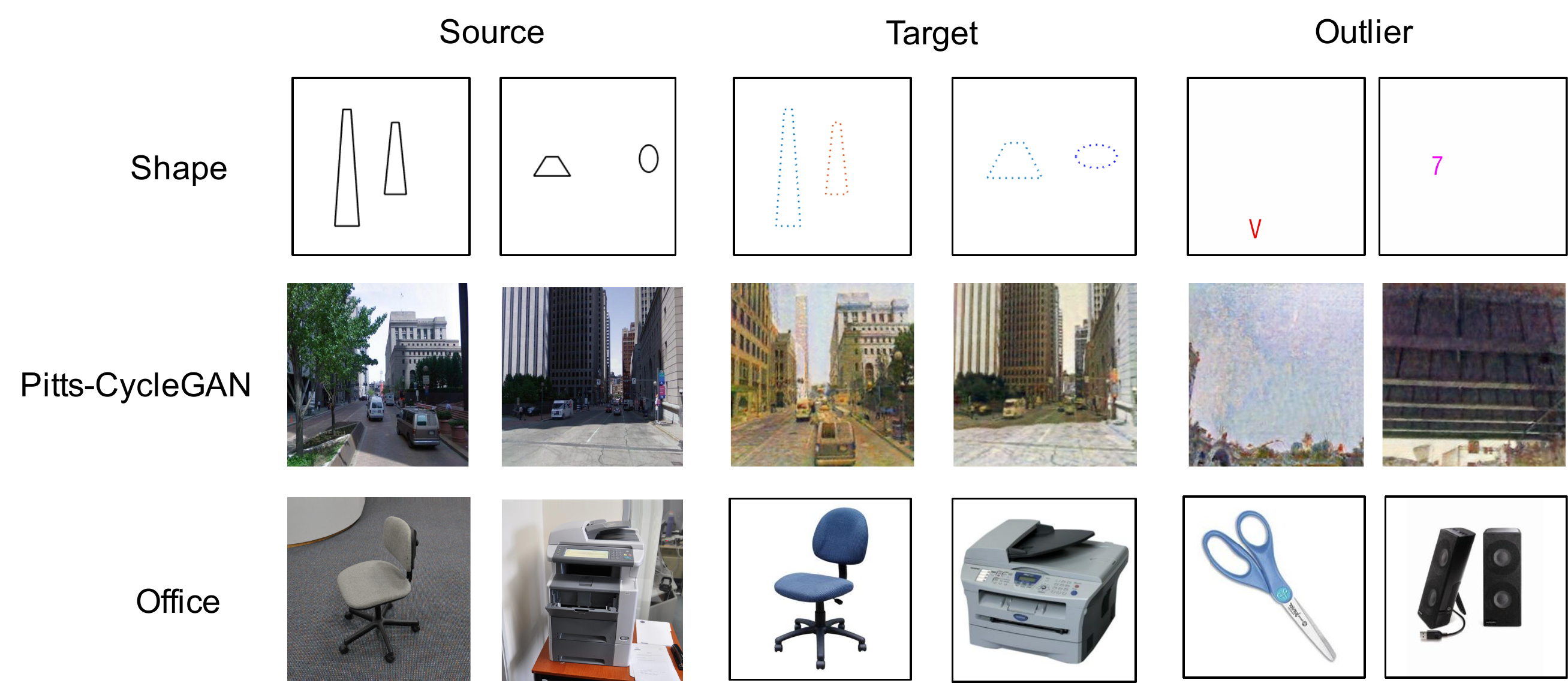}
   \caption{Examples from \textit{Shape}, \textit{Pitts-CycleGAN} and \textit{Office} sets.}
\label{fig:samples}
\end{figure}

There are no publicly available datasets for our task.
Therefore, we propose two datasets for evaluation. 
Sample images from the three datasets are shown in Figure \ref{fig:samples}.

\textbf{Shape} is one of the synthetic datasets we generate. 
It contains 60k source domain images, 30k target domain images (including 2800 outliers).
The outlier images are made up of single alphabets or digits.  
The source domain and inlier images are combinations of two geometric shapes, drawn with black solid lines and colored dot lines, respectively.
We define two images are a matching pair if the combination of shapes is the same.

\textbf{Pitts-CycleGAN} is the other synthetic dataset, which contains 204k Pittsburgh Google Street View images from Pittsburgh dataset~\cite{Torii2013pits} as the source domain, and 157k target domain images (including 52k outliers) generated by applying CycleGAN~\cite{ZhuPIE17CycleGAN} to the Pittsburgh images.
So the target domain images are in a painting style.
The outliers are sky images or city views not containing any useful landmark information.

\textbf{Office}~\cite{Saenko:2010:AVC:1888089.1888106} consists of 3 domains, \textit{Amazon, Dslr, Webcam}.
We choose \textit{Dslr} as source domain and \textit{Amazon} as target domain.
We make pairs with images from the same category.
The outliers come from two randomly chosen categories ('speaker', 'scissors') out of the 31 categories.

\subsection{Implementation details}
\label{sub:5.2}
For our triplet network, the three sub-networks share the same architecture and weights.
Pre-trained AlexNet~\cite{Krizhevsky2012ImageNet} is used for the sub-networks.
We finetune the weights of \textit{conv4 - conv5, fc6, fc7, fc8}. 
For the weighted MK-MMD, we use a Gaussian kernel with a bandwidth $\sigma$ given by the median of the pairwise distances in the training data.
To incorporate the multi-kernel, we vary the bandwidth $\sigma_{m}\in [2^{-8}\sigma, 2^{8}\sigma]$ with multiplicative factor of 2~\cite{Venkateswara2017DHN}.
For performance evaluation, we sort the Euclidean distance between the query and all the gallery features (L2-normalized) to obtain the ranking result.
Moreover, we employ the standard metric mean average precision (MAP).

\subsection{Baseline methods}
\label{sub:5.3}
There are no available baselines to directly compare with our method, thus, we separate our experiments to research on  domain adaptive image matching \ref{sub:5.4} and effectiveness of outlier detection \ref{sub:5.5}.

In the experiment on  domain adaptive image matching, we assume no outliers exist in the target domain. 
Our method is to jointly learn the contrastive loss $L(u_{s})$ and MK-MMD loss $M(u_{s},u_{t})$.
It is trained with pairs from the source domain and images from the target domain, we call it \textbf{\textit{SiameseDA}}.

For evaluating the effectiveness of outlier detection, the target domain contains outliers. 
Our method is called \textbf{\textit{DA+OutlierDetection}}, which learns on the objective \ref{eq:10}.

The baselines for each experiment are shown in Table \ref{table:1}.
\begin{table}[h!]
\begin{center}
\resizebox{\linewidth}{!}{%
\begin{tabular}{l|c}
\hline
Baseline & Experiment\\
[ 0.5ex ]  
 \hline \hline
 & \textbf{ Domain adaptive image matching} \\
 \hline
\textbf{\textit{SIFT + Fisher Vector}}~\cite{Lowe1999SIFT,Sanchez:2013:FisherVector} & trained on the source domain data\\
\textbf{\textit{Siamese}} network~\cite{Chopra2005siamse} & trained on the source domain image pairs\\
 \hline
 & \textbf{Effectiveness of outlier detection} \\
\hline
\textbf{\textit{SiameseDA}} (upper bound) & trained without outliers \\
\textbf{\textit{SiameseDAOut}} (lower bound) & \textbf{\textit{SiameseDA}} trained with outliers\\
 \hline
\end{tabular}}
\end{center}
\caption{Baseline methods for our experiments.}
\label{table:1}
\end{table}

\subsection{Domain adaptive image matching}
\label{sub:5.4}
In this section, we assume the target domain does not contain outliers. We explore if applying domain adaptation improves the performance of cross domain image matching. 
In this case, the learning objective is
\begin{equation}
\label{eq: 11}
min_{u}\textit{J} = \textit{L}(u_{s}) + \gamma\textit{M}(u_{s},u_{t}),
\end{equation}
where, the MK-MMD loss term $\textit{M}(u_{s},u_{t})$ is the unweighted version as explained in subsection \ref{sub:3.2}.
\begin{table*}[t]
\begin{center}
\resizebox{\textwidth}{!}{%
\begin{tabular}{c|c c c|c c c|c c c}
\hline
\multirow{2}{4em}{Method} & \multicolumn{3}{c|}{Shape}&\multicolumn{3}{c|}{Office}&\multicolumn{3}{c}{Pitts-CycleGAN}\\
 & $T\to S$ & $S\to S$ & $T\to T$ &$T\to S$ & $S\to S$ & $T\to T$ &$T\to S$ & $S\to S$ & $T\to T$\\
[ 0.5ex ]  
 \hline \hline
SIFT + Fisher Vector & $2.5\pm0.4$ & $3.6\pm0.3$ &$3.4\pm0.3$ & $3.5\pm0.2$ &$12.0\pm0.5$ & $3.5\pm0.1$& $0.04$ & $0.8\pm0.05$ & $0.3\pm0.03$\\
Siamese & $8.3\pm0.1$ & $95.0\pm0.2$ & $31.7\pm0.6$ & $10.7\pm0.5$ & $99.2\pm0.2$ & $77.2\pm0.3$ & $0.2\pm0.01$ & $81.3\pm0.3$ & $60.6\pm0.5$\\
\hline
\textbf{SiameseDA} & $\textbf{26.4}\pm \textbf{0.2}$ & $53.1\pm0.1$ & $46.2\pm0.1$ & $\textbf{29.1}\pm \textbf{0.1}$ & $99.7\pm0.1$ & $77.5\pm0.2$ & $\textbf{0.4}\pm \textbf{0.01}$ & $80.4\pm0.1$ & $59.5\pm0.1$\\
\hline
\end{tabular}}
\end{center}
\caption{MAP performance for  cross domain image matching and in-domain image matching experiments on three datasets. $T$ means target domain, $S$ means source domain. $T\to S$ implies matching target domain images to source domain images, similar for $S\to S$, $T\to T$. Our method \textbf{SiameseDA} outperforms the baselines across all the datasets.}
\label{table:3}
\end{table*}
\begin{figure*}[t]
\begin{center}
\begin{subfigure}{0.33\textwidth}
\centering
\includegraphics[width=1.0\linewidth]{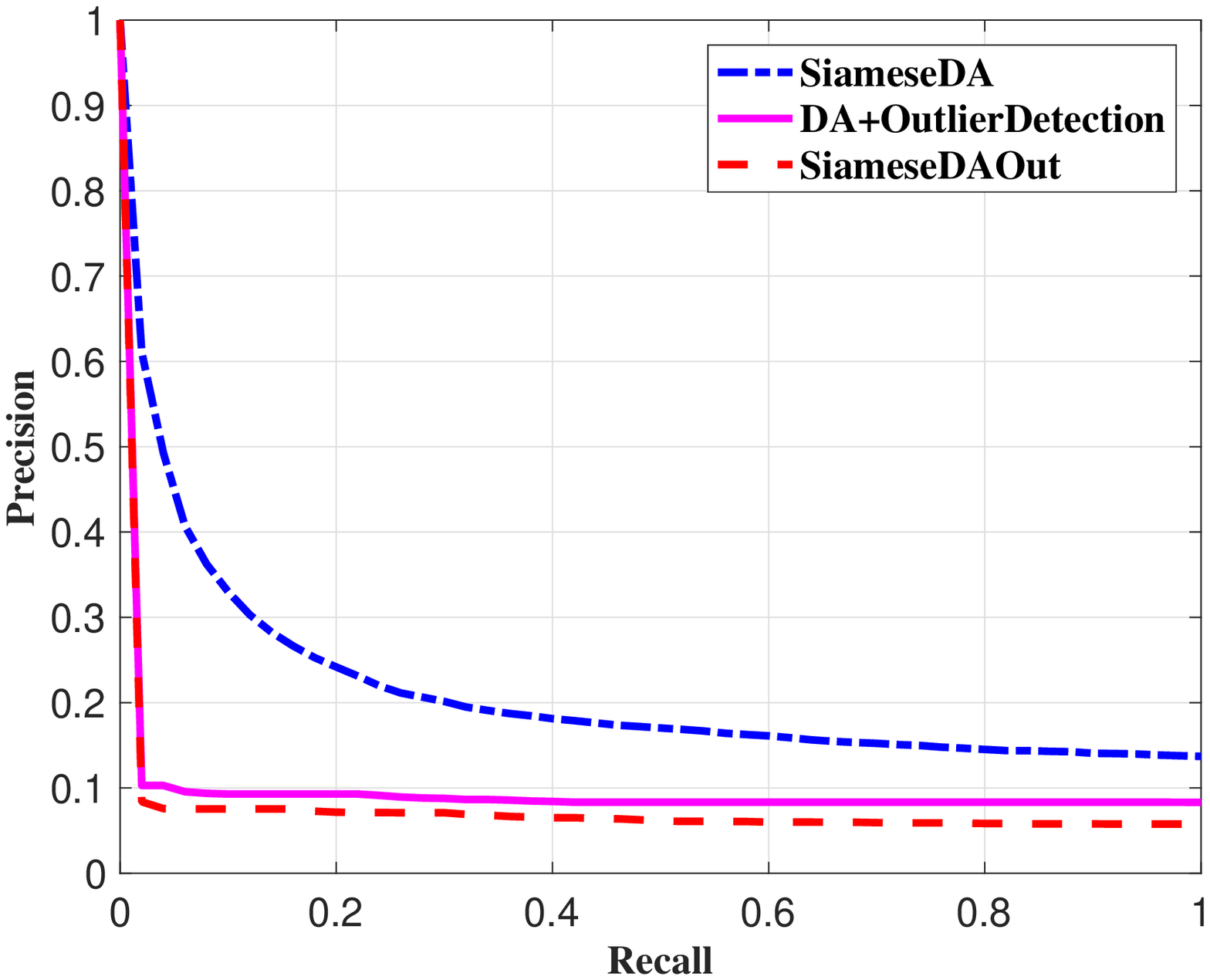}
\caption{Shape}
\end{subfigure}%
\begin{subfigure}{0.33\textwidth}
\centering
\includegraphics[width=1.0\linewidth]{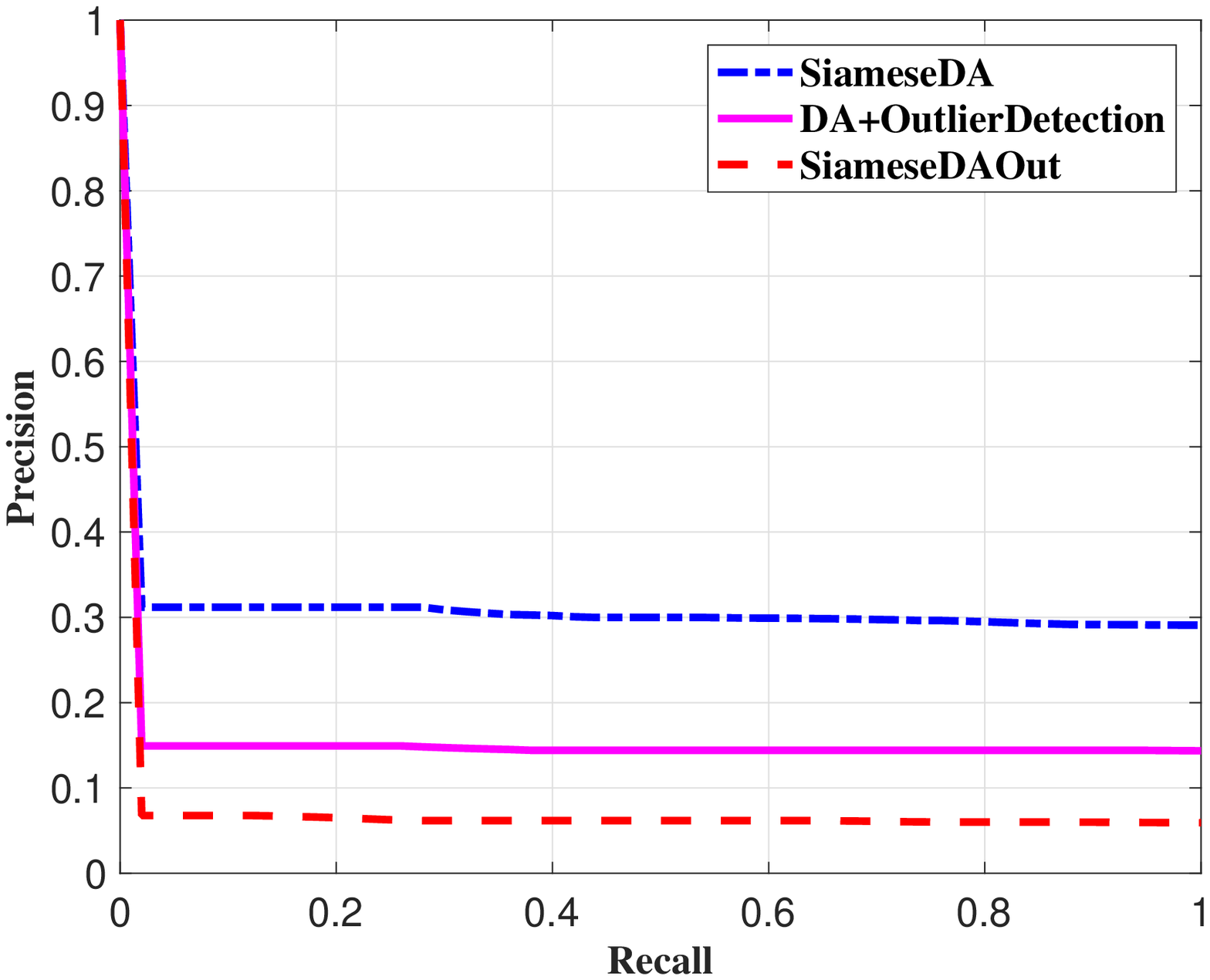}
\caption{Office}
\end{subfigure}%
\begin{subfigure}{0.33\textwidth}
\centering
\includegraphics[width=1.0\linewidth]{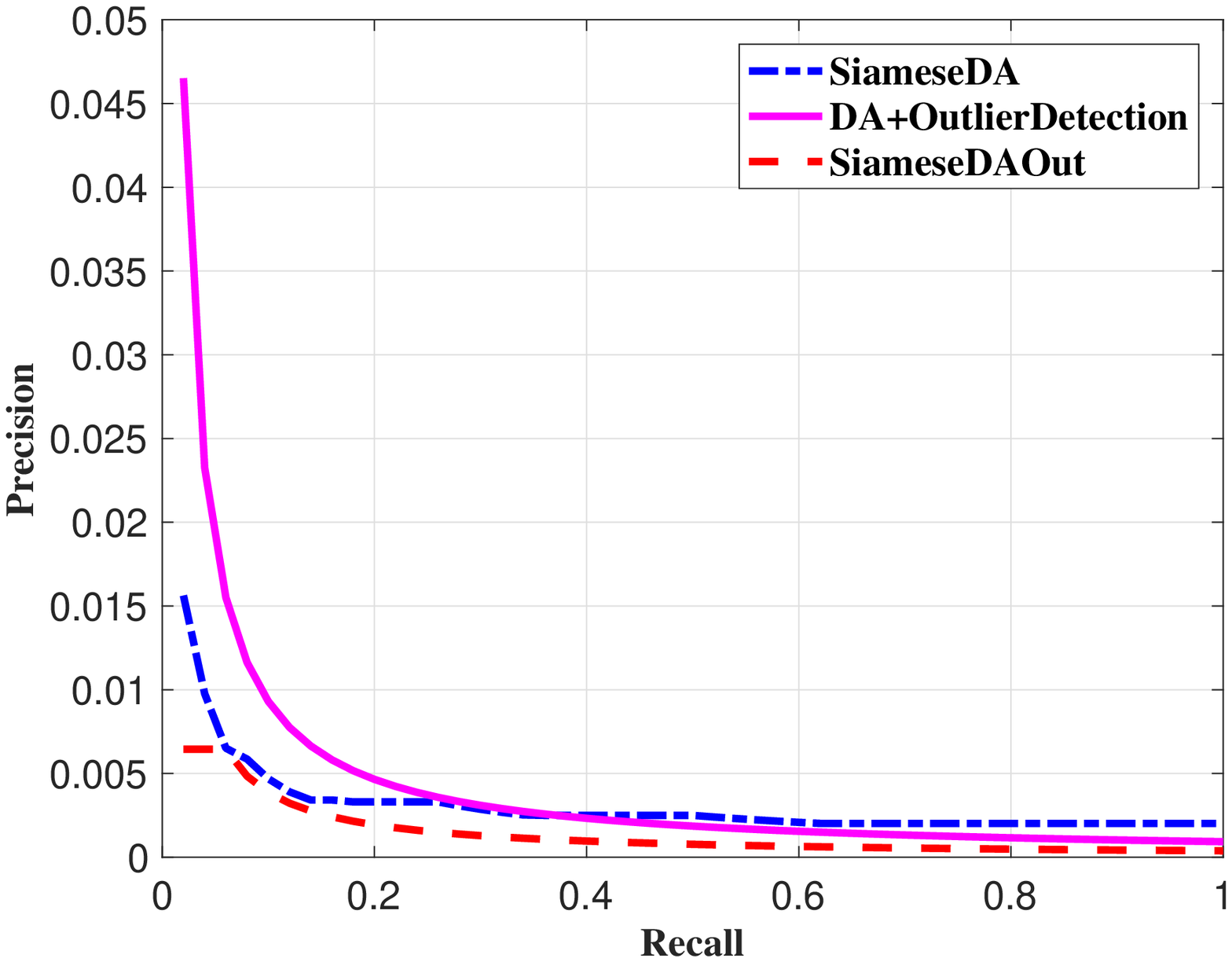}
\caption{Pitts-CycleGAN}
\end{subfigure}
\end{center}
   \caption{Precision-Recall results of our method \textbf{DA+OutlierDetection}, SiameseDA and SiameseDAOut for the experiment of  cross domain image matching with outlier detection on the three datasets. Our method gains over the lower bound method. }
\label{fig:PR2}
\end{figure*}

The MAP results are given in Table \ref{table:3}.
Our method consistently outperforms the baselines across all the datasets.
With applying MK-MMD loss for domain adaptation, the performance of matching $S \to S$ decreases comparing to that of Siamese method.
This is within our expectation since the network may need to learn less from the source domain to be domain adaptive.
Moreover, it is worth to notice that our method also improves the in-domain image matching ($T\to T$) of the target domain.
% % analysis of recall-precision
% \paragraph{The topology}
% With our proposed network architecture and learning objective, we successfully obtain a good result for unsupervised cross domain image matching.
% But we also curious how much the results are contributed by the network architecture or the topology itself.
% Therefore, we have experiments by initializing the network with random weights and no training.
% We test on our three datasets and compare with the training performance we obtain.
% table of results
% \begin{figure*}[h!]
% \centering
% \begin{subfigure}{0.33\textwidth}
% \centering
% \includegraphics[width=1.0\linewidth]{retrToy.png}
% \caption{Shape}
% \end{subfigure}%
% \begin{subfigure}{0.33\textwidth}
% \centering
% \includegraphics[width=1.0\linewidth]{retrOffice.png}
% \caption{Office}
% \label{fig:retrOffice}
% \end{subfigure}%
% \begin{subfigure}{0.33\textwidth}
% \centering
% \includegraphics[width=1.0\linewidth]{retrPitts.png}
% \caption{Pitts-CycleGAN}
% \end{subfigure}
% \caption{Retrieval results for three datasets. For each dataset, the left column shows a query image, the top row shows the top-5
% results for \textit{SiameseDA} method, the middle row shows the top-5 results for our \textit{DA+OutlierDetection} method, and the bottom row shows the top-5 results for \textit{SiameseDAOut} method. Green boxes indicate the corresponding
% correct test impression.}
% \label{fig:retr}
% \end{figure*}

\subsection{Effectiveness of outlier detection}
\label{sub:5.5}
Here we assume the target domain contains outliers, which is to show if the presence of outliers reduces the accuracy of cross domain image matching, and our method could improve it.

The performance of our method (\textit{DA+OutlierDetection}), upper bound (\textit{SiameseDA}) and lower bound (\textit{SiameseDAOut}) are given in Table \ref{table:4}.
In terms of testing, we only take the classified inliers in the query set in calculation.
From Table \ref{table:4} we can see, our method outperforms the lower bound for all the three datasets, but is not better than the upper bound (except for Pitts-CycleGAN) as expected. 
It shows that the presence of outliers reduces the accuracy of cross domain image matching, and our method helps improve the performance in this case.
\begin{table}[h!]
\begin{center}
\resizebox{\linewidth}{!}{%
\begin{tabular}{l|c c c}
\hline
Method ($T \to S$) & Shape & Office & Pitts-CycleGAN\\
[ 0.5ex ]  
 \hline \hline
SiameseDA & $26.4\pm0.2$ & $29.1\pm0.1$ & $0.4\pm0.01$\\
\textbf{DA+OutlierDetection} & $\textbf{11.9}\pm \textbf{0.1}$ & $\textbf{15.9}\pm\textbf{0.2}$& $\textbf{1.1}\pm\textbf{0.03}$\\
SiameseDAOut & $5.4\pm 0.1$ & $6.8\pm0.1$ & $0.2\pm0.01$\\
\hline
\end{tabular}}
\end{center}
\caption{MAP performance for cross domain image matching with outlier detection on our three datasets. The proportion of outliers is 10\%. Our method \textbf{DA+OutlierDetection} outperforms the lower bound, but does not surpass the upper bound.}
\label{table:4}
\end{table}

\begin{figure}[h!]
\centering
\includegraphics[width=0.85\linewidth]{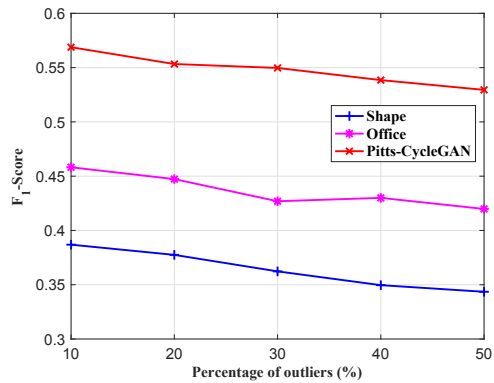}
\caption{$F_{1}$-scores for outlier detection on three datasets with different outlier proportion in the target domain. Our method is consistently robust.}
\label{fig:F1score}
\end{figure}

% precision-recall
In Figure \ref{fig:PR2}, we also show the retrieval performance in terms of the trade-off between precision and recall at different thresholds on our three datasets.
The interpolated average precision is used for the precision-recall curves.
We can see that our method gains over the lower bound method.
\paragraph{Impact of outlier proportion}
We also report the $F_{1}$-score to measure the performance of outlier detection of our method.
Figure \ref{fig:F1score} shows the $F_{1}$-score of our method as a function of the portion of outlier samples for the three datasets.
As can be seen, with the increase in the number of outliers, our method operates consistently robust.

% \paragraph{Data Distribution Change}
% \begin{figure}[h!]
% \centering
% \includegraphics[width=1.0\linewidth]{DisChange.png}
% \caption{The average distance change of target samples to source samples before and after training on our method. The upper subplot is the target sample distribution by distance before training, the bottom one is that after training. The x-axis is the average target sample distance to source data. This is measured on \textit{Shape} dataset.}
% \label{fig:dis}
% \end{figure}
% Since the source training data are the "inlier" reference in our system, we measure the average distance between every testing target sample and source reference data to show the distinguishing ability of our method.
% Figure \ref{fig:dis} shows the distance change of target samples consisting of inliers and outliers before (upper subplot) and after training (bottom subplot) the network on our method with \textit{Shape} dataset.
% The portion of outliers in this experiment is 10\% during training.
% It is obvious that the inliers and outliers in the target domain are still hard to distinguish before training.
% After training the network on our method, we can see that in the bottom subplot, the outliers and inliers are well separated even though it sacrifices some inliers (false negative).

It is important to notice the limitation of our method, which classifies some inlier samples as outliers during training.
This is mainly caused by the way of initializing the probabilities of the target domain training data.
\section{Conclusion}
We have proposed a network that is trained for  cross domain image matching with outlier detection in an end-to-end manner.
The two main parts of our approach are (i) domain adaptive image matching subnetwork with contrastive loss and weighted MK-MMD loss, (ii) outlier detection with entropy loss by updating the probability of target domain data during training.
The results on several datasets demonstrate that the proposed method is capable of detecting outlier samples and achieving  cross domain image matching at the same time.
But our method still needs improvement to overcome the problem of wrongly classifying inliers as outliers.

%-------------------------------------------------------------------------

{\small
\bibliographystyle{ieee}
\bibliography{egpaper_for_review}
}

\end{document}